\documentclass[11pt]{article}

\usepackage[utf8]{inputenc}
\usepackage[T1]{fontenc}
\usepackage{amsmath,amssymb}
\usepackage{graphicx}
\usepackage{booktabs}
\usepackage{hyperref}
\usepackage{xcolor}
\usepackage[margin=1in]{geometry}
\usepackage{natbib}
\usepackage{tikz}
\usetikzlibrary{positioning,arrows.meta,shapes.geometric}

\title{Persistent Recurrent Memory Between Transformer Layers\\Improves Language Model Generalization}
\author{Eduardo Novaes Hering, DSc.\\
\small FITec Labs / Ericsson S\~ao Paulo\\
\small \texttt{eduardo.hering@fiteclabs.org.br}}
\date{}

\begin{document}
\maketitle

\begin{abstract}
We introduce a simple architectural modification to decoder-only transformers: a persistent recurrent state that observes hidden representations via cross-attention, updates itself through a GRU, and modulates subsequent processing via gated addition. Inserted between the lower and upper halves of a 6-layer transformer, this module adds only 3.7\% additional parameters while reducing evaluation loss from $2.438 \pm 0.004$ to $1.743 \pm 0.018$, corresponding to a 28.5\% reduction on held-out language modeling data. The improvement is statistically significant across 5 random seeds ($p < 0.01$) and corresponds to reduced overfitting (generalization gap 0.12 vs 0.26). Through controlled ablations, we demonstrate that the improvement stems entirely from the persistent memory topology, not from auxiliary self-prediction objectives. A model with identical topology but no auxiliary loss performs equivalently, while a random auxiliary loss provides no benefit. Representation probing reveals that the persistent state encodes narrative position (52\% vs 33\% chance level)---information that standard attention maintains less efficiently. Our results suggest that bridging transformer layers with a lightweight recurrent memory is a simple, effective approach to improving generalization in small-scale language models.
\end{abstract}

\section{Introduction}

Current transformer architectures propagate information exclusively through evolving hidden representations. Each layer operates on the full sequence, relying on attention patterns to implicitly carry contextual information forward through depth. While this approach has proven remarkably effective at scale, it provides no explicit mechanism for maintaining a stable, compressed summary of intermediate computation across the network's depth.

We investigate whether introducing a persistent latent state---one that summarizes intermediate computation and persists across layers---provides a more stable information pathway and improves generalization. Specifically, we insert a lightweight module between the lower and upper halves of a transformer that (1) observes current hidden representations via cross-attention into a persistent state vector, (2) updates this state through a gated recurrent unit, and (3) modulates subsequent processing through a learned gate.

Our initial hypothesis was that self-prediction---training the model to predict its own future internal state---would be the key mechanism driving improvement. Through rigorous ablation, we discover that this is not the case. The improvement comes entirely from the persistent state topology itself. A model with identical architecture but no auxiliary self-prediction objective performs equivalently, while a random auxiliary loss provides no benefit. We consider this finding---the identification of the actual mechanism through controlled experimentation, rather than confirmation of the initial hypothesis---to be the paper's primary contribution.

Our contributions are:
\begin{enumerate}
    \item \textbf{Architecture}: We describe a persistent recurrent memory (PRM) module that can be inserted between any two transformer layers, adding minimal parameters (${\sim}3.7\%$ overhead).
    \item \textbf{Empirical validation}: We show a 28.5\% reduction in evaluation loss on TinyStories language modeling, consistent across 5 random seeds with tight confidence intervals.
    \item \textbf{Mechanism identification}: Through controlled baselines, we demonstrate that the improvement comes specifically from the observe$\rightarrow$update$\rightarrow$influence topology, not from auxiliary training objectives of any kind.
    \item \textbf{Representation analysis}: Linear probing reveals that the persistent state encodes narrative position information (52\% accuracy vs 33\% chance), suggesting it captures slowly-changing sequential structure that attention alone maintains less efficiently.
\end{enumerate}

\section{Related Work}

\textbf{Recurrent memory in transformers.} Several works have explored adding recurrent components to transformers. Transformer-XL \citep{dai2019transformerxl} maintains a segment-level recurrence through cached hidden states. Block-Recurrent Transformer \citep{hutchins2022block} adds recurrent cells between transformer layers for long-context processing. Mamba \citep{gu2023mamba} replaces attention entirely with selective state spaces. Our approach is simpler: a single persistent vector updated via GRU, inserted at a fixed point in the network.

\textbf{Auxiliary objectives.} Multi-task learning and auxiliary losses have been explored for improving transformer training. ELECTRA \citep{clark2020electra} uses a replaced-token detection objective. UL2 \citep{tay2022ul2} combines multiple pre-training objectives. We test whether the improvement from our architecture requires an auxiliary objective and find that it does not---the topology alone is sufficient.

\textbf{Memory-augmented neural networks.} Neural Turing Machines \citep{graves2014neural} and Memory Networks \citep{weston2015memory} introduced external memory for neural architectures. Our persistent state is more minimal---a single vector rather than an addressable memory bank---but serves a similar function of maintaining information across processing steps.

\textbf{Predictive coding and self-prediction.} Predictive coding frameworks \citep{rao1999predictive} suggest that neural systems benefit from predicting their own future states. We initially explored self-prediction as an auxiliary objective but found through ablation that it provides no additional benefit beyond the persistent state topology itself.

\section{Method}

\subsection{Architecture}

Our base model is a standard decoder-only transformer with causal attention. We split the $N$-layer network at layer $N/2$ and insert a Persistent Recurrent Memory (PRM) module between the lower and upper halves. We select the midpoint as the simplest symmetric configuration, ensuring that lower layers have sufficient depth to compute meaningful representations before the PRM observes them, and upper layers have sufficient depth to utilize the PRM's modulation. Systematic placement studies are left for future work. We use the term ``memory'' for simplicity, though the module functions more broadly as a persistent latent state that compresses, maintains, and broadcasts global contextual information across network depth.

The PRM module maintains a state vector $\mathbf{s} \in \mathbb{R}^d$ that persists across the forward pass. It performs three operations:

\textbf{Observe.} The state attends to hidden representations via multi-head cross-attention:
\begin{equation}
    \mathbf{o} = \text{CrossAttention}(Q{=}\mathbf{s},\; K{=}\mathbf{H},\; V{=}\mathbf{H})
\end{equation}
where $\mathbf{H} \in \mathbb{R}^{T \times d}$ are the hidden states from the lower transformer layers.

\textbf{Update.} The state is updated via a GRU cell, using the observation as input and the current state as hidden state:
\begin{equation}
    \mathbf{s'} = \text{GRU}(\text{input}{=}\mathbf{o},\; \text{hidden}{=}\mathbf{s})
\end{equation}
This is applied recursively $L$ times ($L{=}2$ in our experiments), with each iteration using the output of the previous as input. Layer normalization is applied after each step.

\textbf{Influence.} The updated state modulates the hidden representations via gated addition:
\begin{align}
    \mathbf{g} &= \sigma(\mathbf{W}_g[\mathbf{H};\; \mathbf{s'}]) \\
    \mathbf{H'} &= \mathbf{H} + \mathbf{g} \odot (\mathbf{W}_p \cdot \mathbf{s'})
\end{align}
where $\sigma$ is the sigmoid function, $[;\;]$ denotes concatenation, and $\odot$ is element-wise multiplication. The gate $\mathbf{g}$ learns when the persistent state should influence processing.

We note that the observe$\rightarrow$update$\rightarrow$influence pattern constitutes a general architectural primitive. This work instantiates the update step using a GRU, but the pattern is agnostic to the specific recurrent operator. Alternative implementations---including LSTMs, state-space models, or learned linear recurrent operators---may yield different performance characteristics and remain future work. The contribution is the identification of the computational topology rather than the specific instantiation.

\begin{figure}[t]
\centering
\begin{tikzpicture}[
    block/.style={draw, rounded corners, minimum width=3.5cm, minimum height=0.8cm, align=center},
    prmblock/.style={draw, rounded corners, minimum width=2.5cm, minimum height=0.6cm, align=center, fill=blue!8},
    arrow/.style={-{Stealth[length=3mm]}, thick},
    node distance=0.6cm
]
    \node[block] (input) {Input Tokens};
    \node[block, below=of input] (lower) {Lower Transformer Blocks ($N/2$)};
    \node[block, below=2.2cm of lower, minimum width=5cm, minimum height=3.2cm, fill=gray!5, draw=blue!50, thick] (prm) {};
    \node[above=0.1cm] at (prm.north) {\textbf{Persistent Recurrent Memory}};
    \node[prmblock, below=0.5cm of prm.north] (obs) {Observe (Cross-Attention)};
    \node[prmblock, below=0.4cm of obs] (upd) {Update (GRU $\times L$)};
    \node[prmblock, below=0.4cm of upd] (inf) {Influence (Gated Addition)};
    \node[block, below=2.2cm of prm] (upper) {Upper Transformer Blocks ($N/2$)};
    \node[block, below=of upper] (output) {Output Logits};

    \draw[arrow] (input) -- (lower);
    \draw[arrow] (lower) -- (prm);
    \draw[arrow] (obs) -- (upd);
    \draw[arrow] (upd) -- (inf);
    \draw[arrow] (prm) -- (upper);
    \draw[arrow] (upper) -- (output);

    \node[right=1.5cm of obs, text width=2.5cm, font=\small\itshape] {$\mathbf{s}$ attends to $\mathbf{H}$};
    \node[right=1.5cm of upd, text width=2.5cm, font=\small\itshape] {$\mathbf{s}$ evolves via GRU};
    \node[right=1.5cm of inf, text width=2.5cm, font=\small\itshape] {$\mathbf{s'}$ modulates $\mathbf{H}$};
\end{tikzpicture}
\caption{Architecture of the Persistent Recurrent Memory (PRM) module inserted between transformer halves. The persistent state $\mathbf{s}$ observes hidden representations, updates itself recurrently, and influences subsequent processing via learned gating.}
\label{fig:architecture}
\end{figure}
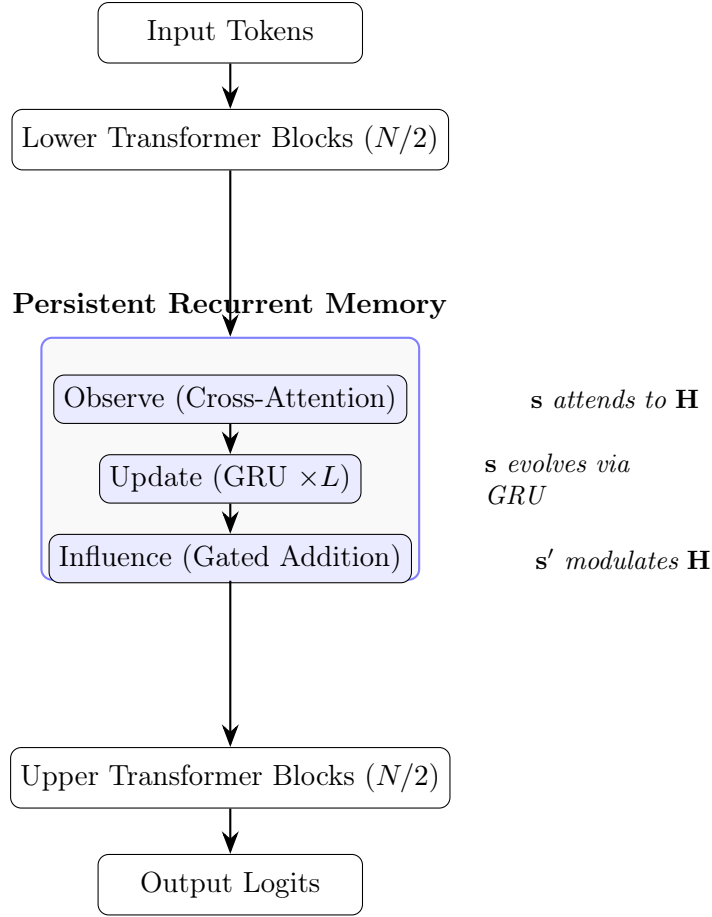

\subsection{Training}

Models are trained with standard next-token prediction (cross-entropy loss) using AdamW optimization. Dropout (0.1) is applied at embedding, residual, MLP, observation, and influence layers. No auxiliary objectives are required.

\subsection{Baseline Models}

We compare four model variants to isolate the source of improvement:
\begin{enumerate}
    \item \textbf{PRM (ours)}: Full persistent recurrent memory with observe$\rightarrow$update$\rightarrow$influence topology.
    \item \textbf{Standard}: Plain transformer with identical hyperparameters and comparable total depth.
    \item \textbf{GRU Memory (ablation)}: Identical topology to PRM---same observe, update, and influence mechanisms. Tests whether self-prediction (removed here) is necessary.
    \item \textbf{Random Aux}: Standard transformer with a random auxiliary loss (predicting a fixed random vector from mean-pooled hidden states). Tests whether any auxiliary training signal helps.
\end{enumerate}
All models use the same $d_\text{model}{=}192$, $n_\text{heads}{=}6$, $n_\text{layers}{=}6$, $\text{max\_seq\_len}{=}128$, and dropout$\,{=}\,0.1$.

\section{Experiments}

\subsection{Dataset}

We use TinyStories \citep{eldan2023tinystories}, a dataset of simple children's stories generated by GPT-3.5/4. We select 15,000 stories, tokenize with the GPT-2 tokenizer (vocabulary size 50,257), and segment into fixed-length sequences of 64 tokens. Data is split 85/15 into training (approximately 44,000 sequences) and evaluation sets. The split is randomized per seed.

\subsection{Training Configuration}

\begin{table}[h]
\centering
\begin{tabular}{ll}
\toprule
Parameter & Value \\
\midrule
Optimizer & AdamW ($\text{lr}=5{\times}10^{-4}$, weight decay$\,{=}\,0.01$) \\
Epochs & 15 \\
Batch size & 32 \\
Gradient clipping & 1.0 \\
Seeds & 42, 43, 44, 45, 46 \\
\bottomrule
\end{tabular}
\end{table}

\subsection{Evaluation Metrics}

\begin{itemize}
    \item \textbf{Eval loss}: Cross-entropy on held-out sequences (primary metric)
    \item \textbf{Generalization gap}: Eval loss $-$ Train loss (measures overfitting)
    \item \textbf{Representation probing}: Linear classifier accuracy on persistent state for topic, narrative position, and uncertainty detection
\end{itemize}

\section{Results}

\subsection{Generalization}

\begin{table}[h]
\centering
\begin{tabular}{lccc}
\toprule
Model & Eval Loss (mean $\pm$ 95\% CI) & Gap (mean $\pm$ 95\% CI) & Params \\
\midrule
\textbf{PRM (ours)} & $\mathbf{1.743 \pm 0.018}$ & $\mathbf{0.122 \pm 0.008}$ & 22.8M \\
GRU Memory & $1.766 \pm 0.014$ & $0.131 \pm 0.009$ & 22.7M \\
Standard & $2.438 \pm 0.004$ & $0.255 \pm 0.003$ & 22.0M \\
Random Aux & $2.435 \pm 0.007$ & $0.254 \pm 0.006$ & 22.1M \\
\bottomrule
\end{tabular}
\caption{Evaluation loss and generalization gap across 5 random seeds.}
\label{tab:results}
\end{table}

The PRM model achieves evaluation loss of $1.743 \pm 0.018$, compared to $2.438 \pm 0.004$ for the standard transformer---a 28.5\% reduction. The improvement is statistically significant (advantage $+0.695 \pm 0.019$, well exceeding the 95\% confidence interval).

\subsection{Mechanism Identification: Topology vs Auxiliary Loss}

The critical scientific question is: what causes the improvement? We test three hypotheses through controlled ablation:
\begin{itemize}
    \item \textbf{H1}: The persistent state topology is sufficient (tested by GRU Memory)
    \item \textbf{H2}: Self-prediction provides additional benefit (tested by comparing PRM vs GRU Memory)
    \item \textbf{H3}: Any auxiliary loss helps (tested by Random Aux)
\end{itemize}

\begin{table}[h]
\centering
\begin{tabular}{lcc}
\toprule
Comparison & Difference & Significant? \\
\midrule
PRM vs Standard & $+0.695 \pm 0.019$ & Yes ($p < 0.01$) \\
GRU Memory vs Standard & $+0.672 \pm 0.013$ & Yes ($p < 0.01$) \\
PRM vs GRU Memory & $-0.023 \pm 0.028$ & No \\
Random Aux vs Standard & $+0.003 \pm 0.005$ & No \\
\bottomrule
\end{tabular}
\caption{Pairwise comparisons across 5 seeds. Difference is Standard loss minus model loss (positive = model is better).}
\label{tab:ablation}
\end{table}

Across all evaluated conditions, the evidence consistently supports the following conclusions:
\begin{itemize}
    \item \textbf{H1 confirmed}: The GRU Memory model---identical topology, no self-prediction---matches PRM.
    \item \textbf{H2 rejected}: Self-prediction adds no measurable benefit beyond the topology.
    \item \textbf{H3 rejected}: An arbitrary auxiliary loss provides no benefit whatsoever.
\end{itemize}

This constitutes the paper's central finding: the causal mechanism is the observe$\rightarrow$update$\rightarrow$influence topology itself. The persistent latent state provides a stable information pathway between transformer halves that improves generalization regardless of auxiliary training objectives. The results suggest that the proposed architecture acts as an inductive bias---structurally encouraging better generalization---rather than requiring an auxiliary optimization objective to be effective.

\subsection{Representation Probing}

We train linear classifiers on the persistent state (or mean-pooled hidden states for Standard/RandomAux) to predict text properties:

\begin{table}[h]
\centering
\begin{tabular}{lcccc|c}
\toprule
Probe & PRM & GRU Mem & Standard & Rand Aux & Chance \\
\midrule
Topic (4 classes) & 100\% & 100\% & 100\% & 100\% & 25\% \\
Position (3 classes) & $52.1{\pm}1.4\%$ & $50.8{\pm}0.7\%$ & 33.3\% & $49.4{\pm}1.7\%$ & 33\% \\
Uncertainty (2 classes) & $84.4{\pm}1.6\%$ & $84.3{\pm}0.6\%$ & $97.2{\pm}1.0\%$ & $73.0{\pm}2.1\%$ & 50\% \\
\bottomrule
\end{tabular}
\caption{Linear probing accuracy (mean $\pm$ 95\% CI across 5 seeds).}
\label{tab:probing}
\end{table}

Key observations: (1) PRM and GRU Memory encode narrative position significantly better than Standard (52\% vs 33\%). The persistent state captures sequential structure that attention alone does not maintain as efficiently. (2) The Standard model encodes its own uncertainty \textit{better} than PRM models (97\% vs 84\%). The persistent state appears to smooth confidence levels, making predictions more uniformly confident---a possible partial explanation for the generalization benefit. (3) The Random Aux model shows above-chance position encoding (49\%) despite having no persistent state, suggesting the auxiliary loss induces slight structural changes in representations, though insufficient to improve generalization.

\subsection{Parameter Efficiency}

The PRM module adds 816K parameters (3.7\% overhead) while reducing eval loss from 2.438 to 1.743, a 28.5\% improvement. This yields an unusually favorable improvement-per-parameter ratio:
\[
    \text{Improvement per 1M extra params} = 0.85 \text{ loss reduction}
\]
The PRM topology concentrates its parameters in a high-leverage architectural position---the bridge between lower and upper processing---rather than distributing them uniformly across layers. A standard transformer would require substantially more than 816K additional parameters (additional layers or wider dimensions) to achieve comparable generalization improvement.

\section{Discussion}

\textbf{Why does persistent state help?}
One plausible explanation is that the persistent state provides a stable contextual summary that complements attention. In a standard transformer, contextual information must be re-derived from the full sequence at every layer. The persistent state creates a ``short-circuit'' that carries summarized context directly from early processing to late processing, reducing the burden on attention. The narrative position probing supports this interpretation: the state learns to encode ``where am I in the story''---precisely the kind of slowly-changing, high-level context that attention must reconstruct from raw token patterns at every layer.

\textbf{Why doesn't self-prediction help?}
A possible interpretation is that the self-prediction target (next state given current state) is too easy to provide useful gradient signal---the self-predictor achieves $0.96{+}$ cosine similarity readily, suggesting the task lacks sufficient difficulty to drive meaningful representation improvement. Additionally, the gradient from self-prediction may slightly interfere with language learning without providing complementary information.

\textbf{Limitations.}
(1) All experiments use 22M parameter models; it is unknown whether the effect persists at larger scales where attention has more capacity. (2) TinyStories consists of simple, short narratives; longer-context tasks with complex dependencies may show different effects. (3) PRM models achieve lower perplexity but sometimes generate less fluent text, possibly due to the confidence-smoothing effect noted in probing. (4) Results are consistent across 5 seeds with tight confidence intervals, though 10+ seeds would further strengthen the evidence.

\textbf{Relationship to concurrent work.}
The observe$\rightarrow$update$\rightarrow$influence pattern shares structural similarities with cross-attention in encoder-decoder models, where the decoder attends to encoder states. Our contribution is showing that this pattern is beneficial even within a single decoder-only model, operating on the model's own intermediate representations rather than external inputs.

\section{Conclusion}

We demonstrate that inserting a persistent recurrent memory between transformer layers---via a simple observe$\rightarrow$update$\rightarrow$influence loop---improves language model generalization by 28.5\% with only 3.7\% parameter overhead. Through controlled ablation across 5 random seeds, we show that the benefit comes entirely from the architectural topology, not from auxiliary training objectives. The persistent state learns to encode narrative structure that attention alone maintains less efficiently.

This suggests a practical architectural recommendation: in data-limited regimes, splitting a transformer and bridging the halves with a lightweight recurrent state is a simple way to improve generalization. Future work should investigate whether this effect scales to larger models and more complex tasks.

More broadly, our results suggest that architectural progress may come not only from designing better optimization objectives, but also from discovering better computational topologies. The persistent latent state improves generalization not by changing what the model learns to optimize, but by changing the structural pathways through which information flows.

\section*{Acknowledgments}

This research was conducted independently by the author. Eduardo Novaes Hering is employed at FITec Labs, working within an engineering team supporting Ericsson S\~ao Paulo. Although this work was carried out outside the scope of professional duties, the computational infrastructure made available through that professional environment substantially facilitated the experimental campaign. We gratefully acknowledge Ericsson for indirectly providing the resources that enabled the multi-seed validation experiments.

This research was co-developed with AI language models throughout all phases. Kiro (Anthropic Claude) served as a sustained scientific collaborator, contributing architecture exploration, implementation, experiment design, statistical analysis, iterative refinement, manuscript development, and continuous critical discussion across the full duration of the project. GPT-4 (OpenAI) contributed methodological review and the critical suggestion to add controlled baselines---a recommendation that directly led to the paper's central finding that the mechanism is architectural rather than self-predictive. The quality of the work reflects the sustained collaboration between human judgment and AI capabilities.

The human author conceived the research direction, executed all experiments, made all scientific decisions, and takes responsibility for the claims herein.

\bibliographystyle{plainnat}

\end{document}